\documentclass[letterpaper]{article}
\usepackage[sort&compress]{natbib}
\usepackage{alifeconf} 

\usepackage[hidelinks]{hyperref}
\usepackage{subfigure}
\usepackage{graphicx}
\usepackage{epstopdf}
\usepackage{url}
\usepackage{breakurl}
\usepackage{mathtools}

\let\OLDthebibliography\thebibliography
\renewcommand\thebibliography[1]{
  \OLDthebibliography{#1}
  \setlength{\parskip}{0pt}
  \setlength{\itemsep}{0pt plus 0.3ex}
}

\title{A Proposed Infrastructure for Adding Online Interaction to Any Evolutionary Domain}
\author{Paul Szerlip \and Kenneth O. Stanley \\
\mbox{}\\
Computer Science Division, University of Central Florida, Orlando, FL 32816 \\
pszerlip@cs.ucf.edu, kstanley@cs.ucf.edu}

\begin{document}
\maketitle

\section{Introduction}

It remains an open problem to effectively leverage the modern Internet's infrastructure to facilitate research in artificial life (ALife) and evolutionary computation (EC). Though inspired by the products of natural evolution, ALife and EC researchers currently operate in relative isolation. A research group building an evolutionary domain, e.g.\ a biped walking domain or a maze navigator, will typically run a series of experiments, publish the final results, provide the code as-is online, and retire the experimental domain. Although the results are published, the ability to reproduce, observe, or extend those results then requires considerable effort outside of the originating group. 

In stark contrast, web technologies provide the means for providing consistently accessible information and uniting distributed communities across the globe. With the rise of cloud computing and advancements in the development of new web technologies, e.g.\ HTML5, JavaScript, and Node.js \citep{dahl:nodejs}, there is an untapped opportunity to create a new set of tools that can enhance the presence of ALife and EC communities online while harnessing the power of humans to aid search on a massive scale.

Interestingly, evolutionary systems can also benefit from many simultaneous users. Collaborate Interactive Evolution (CIE) allows contributions from multiple users over the course of an experiment \citep{szumlanski:geccops05}. In particular, CIE systems like Picbreeder \citep{secretan:ecj11} and Endless Forms \citep{clune:ecal11} accept and benefit from contributions from multiple users online. 

To enhance the potential contribution of each user, these systems explicitly enable users to branch from previously discovered results. For instance, every picture evolved in Picbreeder is either a descendant of another picture inside the system, or started from a simple random starting point. Importantly, branching empowers users of any skill level to contribute to the ongoing experiment while collaboratively building an ever-expanding tree of evolutionary artifacts, i.e.\ a phylogeny. 

In fact, in principle, the power of human intuition can aid evolutionary searches across many domains. For example, recent experiments suggest that humans are capable of interleaving their own insight with automated algorithms, yielding better results than the automated algorithms can alone \citep{bongard:gecco13, woolley:gecco14}.  Yet despite the potential benefits of human-computer collaboration, engineering such a system is still complex and laborious. Picbreeder alone took over a year to build the online infrastructure \citep{secretan:ecj11}. 

To address the difficulty of creating online collaborative evolutionary systems, this paper presents a new prototype library called Worldwide Infrastructure for Neuroevolution (WIN) and its accompanying site WIN Online (\url{http://winark.org/}). The WIN library is a collection of software packages built on top of Node.js that reduce the complexity of creating fully persistent, online, and interactive (or automated) evolutionary platforms around any domain. WIN Online is the public interface for WIN, providing an online collection of domains built with the WIN library that lets novice and expert users browse and meaningfully contribute to ongoing experiments. The long term goal of WIN is to make it trivial to connect any platform to the world, providing both a stream of online users, and archives of data and discoveries for later extension by humans or computers.

 \begin{figure*}[t]
 \begin{center}
   \centering
   \subfigure[win-Picbreeder]{\includegraphics[width=3.2in]{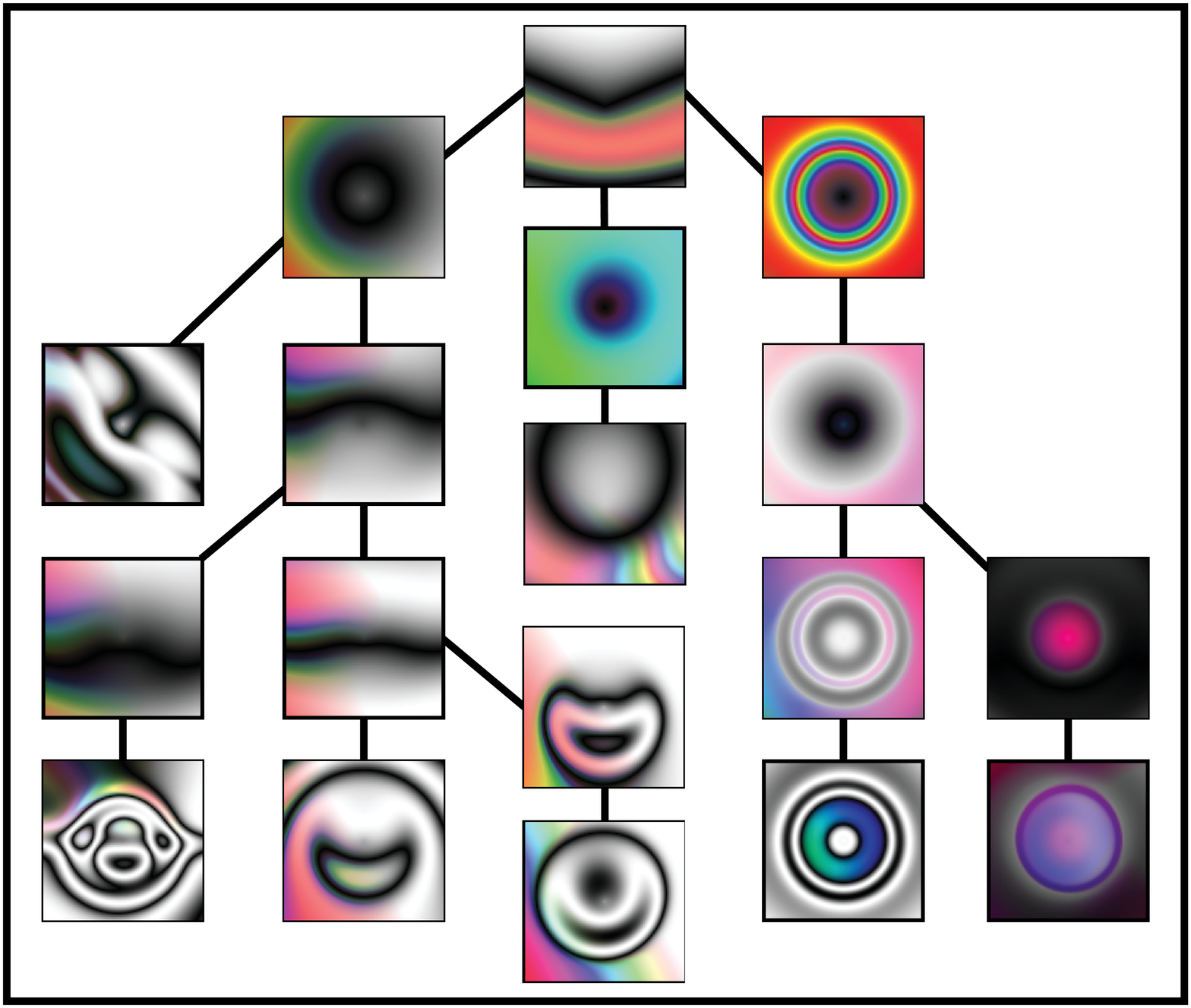}}
   \subfigure[win-IESoR]{\includegraphics[width=3.2in]{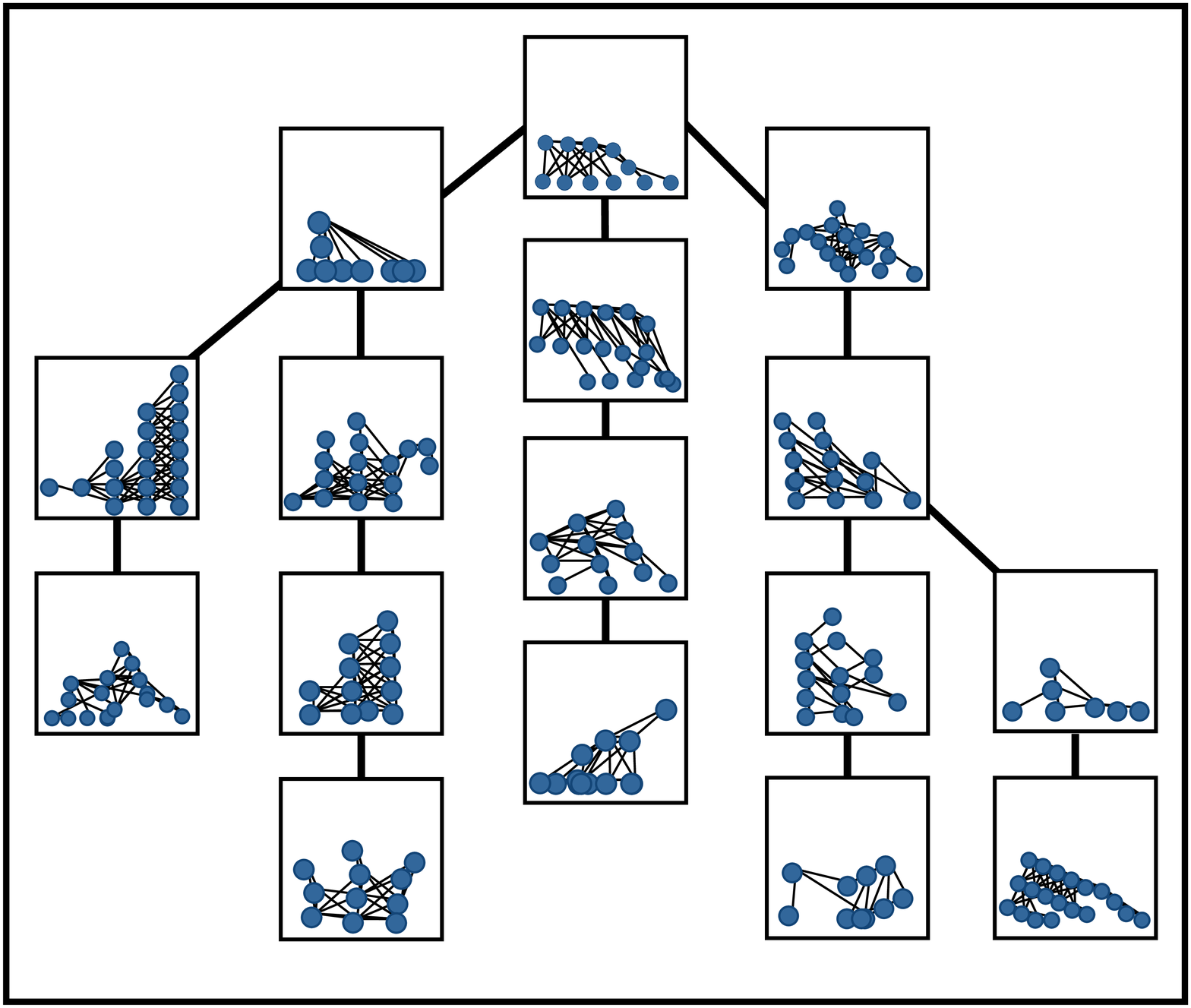}}
 \caption{\textbf{WIN Phylogenies.} 
 The tree of artifacts in (a) and (b) represent artificial phylogenies resulting from the efforts of a single researcher inside two different domains built with WIN. In (a), each square represents a published image inside win-Picbreeder. Each image in (b) illustrates the starting morphology of a two-dimensional ambulating creature evolved by win-IESoR. Both phylogenies can be browsed at \url{http://winark.org/}.
 }
\label{fg:doublePhylogeny}
 \end{center}
 \end{figure*}

\section{Background}

In the fields of ALife and EC, there are continuing efforts to utilize web resources for augmenting ALife research. For example, the ALife Zoo aims to take advantage of cloud computing to host a shared platform for running ALife simulations \citep{hickinbotham:ecal13}. Within evolutionary robotics, \citet{bongard:ludobots} created a simulator, called Ludobots, that sets precedent for harnessing amateur users for crowd-sourcing intricate robot controllers (\url{http://www.uvm.edu/~ludobots/}).  Both works inspire the idea of exploiting advances in web technology to create distributed ALife or EC platforms. WIN builds on the same philosophy with the goal of bringing the benefits of crowd-sourcing to any evolutionary domain. 

\section{WIN Library and WIN Online} 

Built on top of the JavaScript library Node.js \citep{dahl:nodejs}, WIN is a lightweight and expandable collection of Node.js packages that handle storage, retrieval, and cataloging of complex chains of research artifacts. That is, researchers can utilize WIN to quickly create ALife and EC experiments that store and retrieve evolutionary artifacts with a simple web service running on a local machine or across the Internet. 

To handle long-term data storage, WIN employs the NoSQL-database MongoDB. Domains built with the WIN library allow the user to decide when an artifact should be saved for the public record, i.e.\ published. When users publish evolutionary artifacts, the experiment is leveraging human insight to determine when an artifact is of interest. Additionally, WIN provides explicit logic for maintaining relationships among data stored inside the WIN database. As an experiment accumulates data, each object stored by WIN represents a potential branching point for future research with a traceable history of where the data originated. In this way, WIN aims to be a stepping stone accumulator for any evolutionary domain, allowing previously-finite experiments to exploit the possibilities of never-ending search. To illustrate the potential of a WIN experiment, figure \ref{fg:doublePhylogeny} shows example phylogenies built from evolutionary data stored in the WIN database of two such experiments. 

An exciting aspect of WIN is the ability to create a wide range of experiments that can be linked to a common platform, WIN Online. Hosted at \url{http://winark.org/}, WIN Online is a worldwide repository of ongoing experiments for any evolutionary or artificial life domain integrated with the WIN platform. Thus WIN Online becomes a place where users who are interested in the ALife and EC communities can participate in academic domains without prerequisite domain knowledge and immediately start a fresh evolutionary branch from existing results. 

\section{Conclusion}

WIN is a new ALife and EC tool that gives researchers the ability to quickly make their research available to users online, who can then genuinely contribute. Furthermore, WIN Online aggregates domains built with the WIN platform into a single source for new and exciting research that allows users to browse and participate.

\section*{Acknowledgments}
This work was supported by the National Science Foundation under grant no. IIS-1002507. Any opinion, findings, and conclusions or recommendations expressed in this material are those of the authors(s) and do not necessarily reflect the views of the National Science Foundation.

\footnotesize



\bibliographystyle{apalike}
\bibliography{ucf,nn,novelty,nsf_newcites,win}

\begin{thebibliography}{}

\bibitem[Bongard, 2013]{bongard:ludobots}
Bongard, J.~C. (2013).
\newblock Ludobots website.
\newblock The Ludobots software is publicly available at
  http://www.uvm.edu/~ludobots/.

\bibitem[Bongard and Hornby, 2013]{bongard:gecco13}
Bongard, J.~C. and Hornby, G.~S. (2013).
\newblock Combining fitness-based search and user modeling in evolutionary
  robotics.
\newblock In {\em Proceeding of the Fifteenth Annual Conference on Genetic and
  Evolutionary Computation Conference}, GECCO '13, pages 159--166, New York,
  NY, USA. ACM.

\bibitem[Clune and Lipson, 2011]{clune:ecal11}
Clune, J. and Lipson, H. (2011).
\newblock Evolving three-dimensional objects with a generative encoding
  inspired by developmental biology.
\newblock In {\em Proceedings of the European Conference on Artificial Life
  ({ECAL}-2011)}, pages 141--148.

\bibitem[Dahl, 2009]{dahl:nodejs}
Dahl, R.~L. (2009).
\newblock Node.js software package.
\newblock The Node.js software package is publicly available at
  http://nodejs.org/.

\bibitem[Hickinbotham et~al., 2013]{hickinbotham:ecal13}
Hickinbotham, S., Weeks, M., and Austin, J. (2013).
\newblock The alife zoo: cross-browser, platform-agnostic hosting of artificial
  life simulations.
\newblock In {\em Proceedings of the European Conference on Artificial Life
  (ECAL-2013)}.

\bibitem[Secretan et~al., 2011]{secretan:ecj11}
Secretan, J., Beato, N., D.Ambrosio, D.~B., Rodriguez, A., Campbell, A.,
  Folsom-Kovarik, J.~T., and Stanley, K.~O. (2011).
\newblock Picbreeder: A case study in collaborative evolutionary exploration of
  design space.
\newblock {\em Evolutionary Computation}, 19(3):345--371.

\bibitem[Szumlanski et~al., 2005]{szumlanski:geccops05}
Szumlanski, S.~R., Wu, A.~S., and Hughes, C.~E. (2005).
\newblock Collaborative interactive evolution.
\newblock In {\em Proceedings of the Genetic and Evolutionary Computation
  Conference (GECCO) Poster Session}, New York, NY. ACM Press.

\bibitem[Woolley and Stanley, 2014]{woolley:gecco14}
Woolley, B.~G. and Stanley, K.~O. (2014).
\newblock Novel human-computer collaboration: Combining novelty search with
  interactive evolution.
\newblock In {\em Proceedings of the Genetic and Evolutionary Computation
  Conference ({GECCO}-2014)}, New York, NY, USA. ACM.
\newblock To appear.

\end{thebibliography}


\end{document}